%% file: main.tex
\documentclass{article}
\usepackage{iftex}
\ifPDFTeX
  \usepackage[utf8]{inputenc}
\fi
\usepackage[T1]{fontenc}
\usepackage{genai4health_2026}
\usepackage{hyperref}
\usepackage{url}
\usepackage{booktabs}
\usepackage{amsmath,amssymb}
\usepackage{graphicx}
\hypersetup{hidelinks}

\usepackage{tabularx}
\usepackage{graphicx}
\usepackage{fancyvrb}
\usepackage{fvextra}
\usepackage{array}

\newcommand{\sysname}{VitalAgent}
\newcommand{\dataname}{VitalBench}

\title{VitalAgent: A Tool-Augmented Agent for Reactive and Proactive Physiological Monitoring over Wearable Health Data}
\author{%
  \makebox[0.45\textwidth]{Di Zhu} \\
  The University of Melbourne \\
  Melbourne, Australia \\
  \texttt{di.zhu3@student.unimelb.edu.au}
  \And
  \makebox[0.45\textwidth]{Yu Yvonne Wu} \\
  Dartmouth College \\
  Hanover, NH, USA \\
  \texttt{yvonne.wu@dartmouth.edu}
  \And
  \makebox[0.45\textwidth]{Hong Jia} \\
  University of Auckland \\
  Auckland, New Zealand \\
  \texttt{hong.jia@auckland.ac.nz}
  \And
  \makebox[0.45\textwidth]{Aaqib Saeed} \\
  Eindhoven University of Technology \\
  Eindhoven, Netherlands \\
  \texttt{a.saeed@tue.nl}
  \And
  \makebox[0.45\textwidth]{Vassilis Kostakos} \\
  The University of Melbourne \\
  Melbourne, Australia \\
  \texttt{vassilis.kostakos@unimelb.edu.au}
  \And
  \makebox[0.45\textwidth]{Ting Dang} \\
  The University of Melbourne \\
  Melbourne, Australia \\
  \texttt{ting.dang@unimelb.edu.au}
}

\begin{document}
\maketitle

\begin{abstract}
Wearable devices provide continuous access to physiological signals such as ECG and PPG, offering a path to answering users' questions about their own physiological state and to early detection of cardiac and physiological abnormalities. Yet existing mobile health (mHealth) systems are largely limited to task-specific prediction pipelines or reactive question answering over static summaries. They lack the temporal reasoning, persistent physiological context, and proactive monitoring required for long-term signal streams. To advance this setting, we introduce \textbf{\dataname{}}, a longitudinal physiological monitoring benchmark comprising 1,862 QA pairs for reactive question answering and 90.2 hours of continuous ECG/PPG recordings for proactive monitoring, covering cardiac, physical activity, and stress-related tasks. Building on this resource, we propose \textbf{\sysname{}}, a tool-augmented agentic framework for ECG/PPG-based mHealth that supports both reactive question answering and proactive monitoring. \sysname{} is built on a longitudinal physiological memory and a tool-augmented reasoning interface that enables dynamic computation over raw signals. Experiments demonstrate that \sysname{} achieves over 25 points absolute improvement over prompt-based and agentic baselines in reactive evaluation and enables proactive alert monitoring over long-term physiological signals, pointing toward mHealth systems that can both answer questions about and continuously monitor a person's physiological state beyond the clinic.
\end{abstract}

\section{Introduction}
\input{sections/01_intro}

\section{Related work}
\input{sections/02_rw}

\section{\dataname{} benchmark}
\input{sections/03_QA}

\section{Methods}
\input{sections/04_method}

\section{Experiment setup}
\input{sections/05_experiments}

% Results
\input{sections/06_results}

\section{Conclusion}
\input{sections/07_conclusion}

\bibliographystyle{plainnat}
\bibliography{references}

% Appendix
\input{sections/08_appendix}

\end{document}

%% file: sections/01_intro.tex
Cardiac conditions such as atrial fibrillation (AF) are common, often intermittent, and frequently undiagnosed until a serious event occurs. Detecting such conditions often requires chronic monitoring, which traditionally relies on specialized medical-grade equipment and can be difficult to sustain in everyday life. This limitation has motivated the growing use of wearable devices in mHealth, which enable continuous monitoring of physiological signals in daily life, including motion, respiration, sleep, and electrodermal activity~\citep{gomes2023survey}.

Among these, cardiac sensing modalities such as electrocardiography (ECG) and photoplethysmography (PPG) are the most widely used in consumer and clinical wearables, providing longitudinal views of heart rhythm, variability, recovery, and cardiovascular dynamics outside clinical settings~\citep{de2019wearable,hickey2021smart}. These capabilities enable new forms of continuous and personalized health monitoring. Population-scale studies have shown that such devices can help identify previously undiagnosed AF outside clinical settings~\citep{NEJMoa1901183, guo2019mobile}. At the same time, this shift places continuous physiological data directly in the hands of non-clinicians, creating a need to translate these data into clear, accessible, and actionable health insights.

Despite this progress, current mHealth systems remain limited in how they transform continuous physiological streams into actionable insights. Most existing approaches rely on task-specific pipelines that map physiological signals to predefined outputs such as stress, sleep stages, arrhythmia detection, or activity recognition~\citep{boe2019automating,vos2023generalizable}. While effective for narrow tasks, they struggle with natural language queries, such as “Has my resting heart rate increased recently?” or “How often was my rhythm irregular in the past hour?”, which require retrieval, temporal comparison,  aggregation over longitudinal physiological history, and natural interactions. 

Recent large language model (LLM)-based health assistants improve interaction flexibility through natural-language interaction~\citep{medpalm, amie}, but they typically operate over textual summaries or engineered prompts of physiological signals~\citep{phllm, healthllm}. The underlying physiological evidence is usually generated offline and continuously. 
However, different user questions require different forms of physiological analysis, ranging from instantaneous measurements (e.g., heart rate), to rhythm characterization (e.g., AF detection), and long-term aggregation (e.g., AF burden). Static summaries therefore provide limited support for query-specific reasoning. Without direct access to raw physiological signals and dynamic signal-analysis capabilities, these systems cannot reliably compute personalized evidence on demand, or maintain persistent monitoring context across interactions~\citep{physiologicaldataanalysis}. 

To support these user-facing health queries in daily health monitoring, we introduce \textbf{\dataname{}}, the first longitudinal physiological benchmark dataset unifying ECG and PPG data for both reactive and proactive evaluation, spanning cardiac, activity and stress. Built from four public datasets of long-duration annotated recordings, it pairs user-centric questions with deterministically grounded answers, covering both window-local queries and multi-window queries requiring temporal aggregation. For proactive evaluation, it further provides 90.2 hours of continuous recordings with time-aligned rhythm annotations, allowing proactive alerting to be assessed against annotated abnormal episodes.

Building on this dataset, we introduce \textbf{\sysname{}}, a unified agentic mHealth framework for ECG- and PPG-based physiological monitoring that supports both reactive question answering and proactive monitoring. \sysname{} combines dynamic tool-augmented reasoning with a longitudinal physiological memory. It integrates 29 modular mHealth tools that enable query- and state-dependent analysis of raw physiological signals, while maintaining a physiological memory that stores historical signals. 
Extensive experiments demonstrate that \sysname{} substantially outperforms existing wearable-health pipelines in reactive question answering, highlighting the benefit of dynamic tool augmentation for wearable health queries. It also demonstrates the potential of the framework for proactive monitoring with guideline-informed health alerts.

This design also has practical implications for privacy-aware deployment. Physiological computation is externalized into modular tools, and the LLM only orchestrates, never accessing the raw physiological data directly. This design therefore maps naturally onto emerging mHealth Model Context Protocol (MCP)-style architectures, preserving privacy while maintaining modularity. Our contributions are summarized as follows:
\begin{itemize}

\item We introduce \textbf{\dataname{}}, a longitudinal benchmark dataset comprising 1,862 QA pairs for reactive questions and 90.2 hours of data for proactive monitoring.

\item We propose \textbf{\sysname{}}, a novel and unified agentic mHealth framework supporting reactive QA and proactive monitoring via shared memory and tools.

\item We demonstrate that \sysname{} outperforms reactive baselines by over 25 points and enables proactive alerting for continuous monitoring. 
\end{itemize}
This work paves the way for wearable physiological monitoring more broadly, enabling flexible, agentic systems that integrate continuous sensing with dynamic reasoning. 

%% file: sections/02_rw.tex
\paragraph{Foundation models for mobile health. }
Foundation models have recently been explored for physiological signals, including PPG representation learning~\citep{normwear,papagei} and ECG foundation models~\citep{ecgfounder}. However, they mainly focus on prediction or representation learning, rather than interactive question answering over continuous physiological signals.

\paragraph{LLM-based and agentic mobile‑health models.} 
Recent work has explored the integration of LLMs into wearable health monitoring pipelines. Health-LLM~\citep{healthllm} and PH-LLM~\citep{phllm} adapt LLMs to wearable health prediction and personalized health tasks; however, both approaches rely on precomputed physiological signal statistics passed to the LLM as text summaries. These statistics are predefined and query-agnostic: they remain fixed regardless of what the user asks, even when answering different questions may require examining fundamentally different signal properties. 

Agentic systems further add planning, tool use, retrieval, and iterative computation~\citep{phia, lifeagentbench,imran2024llasa}, but they differ fundamentally from \sysname{} in three aspects: they primarily i) reason over behavioral health data (e.g., activity, sleep, and wearable summaries) rather than continuous physiological signals, ii) rely on tools for information retrieval and statistical analysis rather than physiological signal interpretation, and iii) function as reactive assistants that respond to user queries instead of proactively monitoring health states and initiating interventions.

\paragraph{mHealth QA datasets.}
To support personal health assistants, a growing body of work pairs physiological or wearable signals with natural-language questions to evaluate whether models can ground user queries in signal evidence for response. ECG-QA and related datasets provide template-based, expert-grounded reasoning over electrocardiograms~\citep{ecgqa,ecgexpertqa,fslecgqa}, PulseLM formulates PPG understanding as large-scale QA over pulse segments~\citep{pulselm}, and broader sensor-QA benchmarks such as OpenSQA and SensorQA study language-grounded reasoning over inertial and daily-life signals~\citep{imran2024llasa,sensorqa}. While valuable, these benchmarks are typically limited to i) single modalities, ii) short single-window queries, and iii) mainly clinical QAs but less user-facing question framing, which diverges from how real users interact with wearables. In contrast, our benchmark emphasizes \emph{user-centric phrasing in the mHealth context} and supports both reactive and proactive monitoring.

%% file: sections/03_QA.tex
\label{sec:benchmark}
We introduce \dataname{}, the first longitudinal physiological benchmark with natural-language queries that unifies ECG and PPG for both reactive and proactive evaluation.

\subsection{Dataset sources and windowing}
Different from existing physiological QA benchmarks that focus on short-term prediction, we construct \dataname{} from four public datasets containing long-duration physiological recordings with reference annotations or protocol labels. As summarized in Table~\ref{tab:datasets}, these include: Icentia11k~\citep{icentia11k, pollard2026physionet, tan2019icentia11k} (long-term single-lead ECG with rhythm annotations), AF-PPG-ECG~\citep{afppgecg} (wrist PPG with ECG-derived AF labels), PPG-DaLiA~\citep{ppg-dalia} (wrist PPG with activity labels and HR), and WESAD~\citep{wesad} (multimodal stress labels), covering three health domains. 

Continuous recordings are segmented into ordered monitoring windows following dataset-specific conventions rather than a fixed global length: 5-minute windows for Icentia11k and AF-PPG-ECG, and 30--60 second windows for PPG-DaLiA and WESAD. Window length follows the timescale at which each phenomenon is annotated, so that every window can be paired with a reference label.

\begin{table}[t]
\centering
\footnotesize
\setlength{\tabcolsep}{4pt}
\caption{Source datasets used to construct \dataname{}, covering ECG and PPG across three domains.}
\begin{tabular}{cllll}
\toprule
\# & Dataset & Domain & Signal & Annotations \\
\midrule
1 & Icentia11k
& Cardiac
& ECG
& Rhythm labels \\

2 & AF-PPG-ECG
& Cardiac
& PPG
& AF labels \\

3 & PPG-DaLiA
& Activity
& PPG
& Activity, HR \\

4 & WESAD
& Stress
& PPG
& Stress labels \\

\bottomrule
\end{tabular}
\label{tab:datasets}
\end{table}

\subsection{Reactive QA data construction} 
\label{sec:reactiveQA}
\paragraph{Question templates.}
Questions are designed to reflect how general users ask about wearable data, rather than clinical expert queries. We define targets from dataset annotations, such as AF presence, or signal-derived statistics, such as heart rate. For each target, an LLM generates one canonical question and two paraphrases for QA construction.

We define two question tiers:

\begin{itemize}
\item Tier A: window-local questions over the current window, requiring instantaneous reasoning from the current physiological state, such as “Is there AF right now?” or “What is my current heart rate?”.
\item Tier B: multi-window questions requiring temporal aggregation, comparison, or trend computation across a monitoring period, such as “What was the AF burden in the past hour?” or “What was the maximum heart rate?”.
\end{itemize}

For both tiers, the LLM is prompted to generate concise, first-person, user-facing questions, with the template given in Table~\ref{tab:prompt}. The \texttt{answer\_set} defines the response space (yes/no, multiple-choice, or numeric questions), while \texttt{required\_fields} specifies the target physiological signal, such as \texttt{af\_burden\_ratio},  \texttt{max\_hr\_bpm}, or \texttt{stress\_label}.

\begin{table}[t]
\centering
\footnotesize
\caption{Prompt for generating user-facing questions. Examples in parentheses illustrate each field.}
\begin{tabular}{p{0.95\linewidth}}
\toprule
You are helping design natural-language questions for a mobile health QA benchmark. Generate only natural user questions in a concise, first-person style. Do not include answer-format instructions. \\
\midrule
\textbf{target:} \{target\} \hfill \textit{(e.g., af\_presence\_current)} \\
\textbf{question\_type:} \{question\_type\} \hfill \textit{(e.g., single\_verify)} \\
\textbf{time\_scope:} \{time\_scope\} \hfill \textit{(e.g., current\_window)} \\
\textbf{required\_fields:} \{required\_fields\} \hfill \textit{(e.g., rhythm\_class)} \\
\textbf{answer\_set:} \{answer\_set\} \hfill \textit{(e.g., "yes", "no")} \\
\midrule
\textbf{Output:} 1 canonical question and 2 paraphrases. \\
\bottomrule
\end{tabular}
\label{tab:prompt}
\end{table}

\paragraph{QA pairing. }
Once questions are generated, we construct answer pairs grounded in annotations or signal-derived parameters. For example, we pre-compute target variables (e.g., heart rate, AF presence) for each window in Tier A and pair them with the corresponding questions. Since all answers are either directly provided or computed via deterministic rule-based extraction, answers are strictly grounded and ensure factual consistency. 

For Tier B, the process is analogous, except that answers are derived from aggregated or temporal fields (e.g., maximum, mean, or burden statistics) computed over multi-window states. 

\paragraph{QA types. }We restrict to three QA types: i) \texttt{single\_verify}, referring to binary verification questions with yes/no answers;
ii) \texttt{single\_choose}, referring to categorical selection questions where the answer is chosen from predefined options;
and iii) \texttt{single\_query}, referring to direct querying questions whose answer is a short physiological value or label. 
Two example QA pairs are shown in Table~\ref{tab:qa_example}.

The final benchmark comprises 1,862 QA pairs in total as shown in Table~\ref{tab:benchmark_stats}, with 906 Tier A (window‑local) and 956 Tier B (multi-window) examples. They are stratified by dataset and difficulty tier (Appendix~\ref{app:dataset_details}).

\begin{table}[t]
\centering
\small
\setlength{\tabcolsep}{6pt}
\caption{Example QA pairs from Tier A and Tier B in \dataname{}.}
\begin{tabular}{p{0.46\linewidth}p{0.46\linewidth}}
\toprule
\textbf{Tier A} & \textbf{Tier B} \\
\midrule

\textbf{Question:} \emph{Does my recording indicate atrial fibrillation right now?}
&
\textbf{Question:} \emph{How often did my recent data show AF: not at all, occasionally, often, or most of the time?}
\\[0.6em]

\textbf{Answer:} \emph{no}
&
\textbf{Answer:} \emph{occasionally}
\\[0.4em]

\textbf{Question type:} \emph{single\_verify}
&
\textbf{Question type:} \emph{single\_choose}
\\

\bottomrule
\end{tabular}
\label{tab:qa_example}
\end{table}

\begin{table}[t]
\centering
\footnotesize
\setlength{\tabcolsep}{3.5pt}
\caption{QA pair distribution. V/C/Q denote verify, choose, and query.}
\begin{tabular}{lrrrrrrr}
\toprule
            & \multicolumn{3}{c}{Tier A} & \multicolumn{3}{c}{Tier B} & \\
\cmidrule(lr){2-4} \cmidrule(lr){5-7}
Dataset     & V & C & Q & V & C & Q & Total \\
\midrule
Icentia11k  & 167 & 56  & 55  & 112 & 111 & 55  & 556 \\
AF-PPG-ECG  & 167 & 56  & 55  & 112 & 111 & 55  & 556 \\
PPG-DaLiA   & 67  & 67  & 66  & 100 & 0   & 100 & 400 \\
WESAD       & 100 & 50  & 0   & 50  & 50  & 100 & 350 \\
\midrule
\textbf{Total} & \textbf{501} & \textbf{229} & \textbf{176} & \textbf{374} & \textbf{272} & \textbf{310} & \textbf{1862} \\
\bottomrule
\end{tabular}
\label{tab:benchmark_stats}
\end{table}

\subsection{Proactive monitoring data construction}

Unlike the QA portion of \dataname{}, the proactive monitoring subset is formulated as longitudinal physiological streams with time-stamped monitoring events. For each window, we extract signal-derived measurements, such as heart rate, HRV, rhythm classification, and temporal trends, and align them with reference annotations to obtain continuous physiological trajectories and rhythm transitions.
We select Icentia11k~\citep{icentia11k, pollard2026physionet, tan2019icentia11k} (25 participants) and AF-PPG-ECG~\citep{afppgecg} (44 participants) for proactive monitoring evaluation because their fine-grained, time-aligned rhythm/AF annotations capture AF-related rhythm transitions. These annotations can be aligned with streaming monitoring windows to support alert detection, while the long-duration recordings approximate continuous monitoring. 

The system generates alerts for AF, and we use the annotated events as ground truth for evaluation. Alerts issued within an AF episode count toward that episode, while those issued in reference non-AF windows are counted as unmatched alerts. The two datasets collectively cover 69 participants and 90.2 hours of monitoring data. Recordings are presented as a stream of non-overlapping 10-second windows.  To avoid counting repeated alerts within one abnormal event, alert coverage is evaluated per episode: an episode counts as alerted once, however many alerts it receives. Each episode is defined as a contiguous sequence of windows sharing the same normalized abnormal rhythm label, representing a continuous period of abnormal physiological activity. Of the 32,475 windows, 1,787 were labeled as AF (AFIB and AFL were combined into AF), constituting 47 reference AF episodes, of which 40 were AFIB and 7 were AFL. This class imbalance reflects the natural population prevalence.

%% file: sections/04_method.tex
\subsection{Overview}
\begin{figure*}[t]
    \centering
    \includegraphics[width=0.9\textwidth]{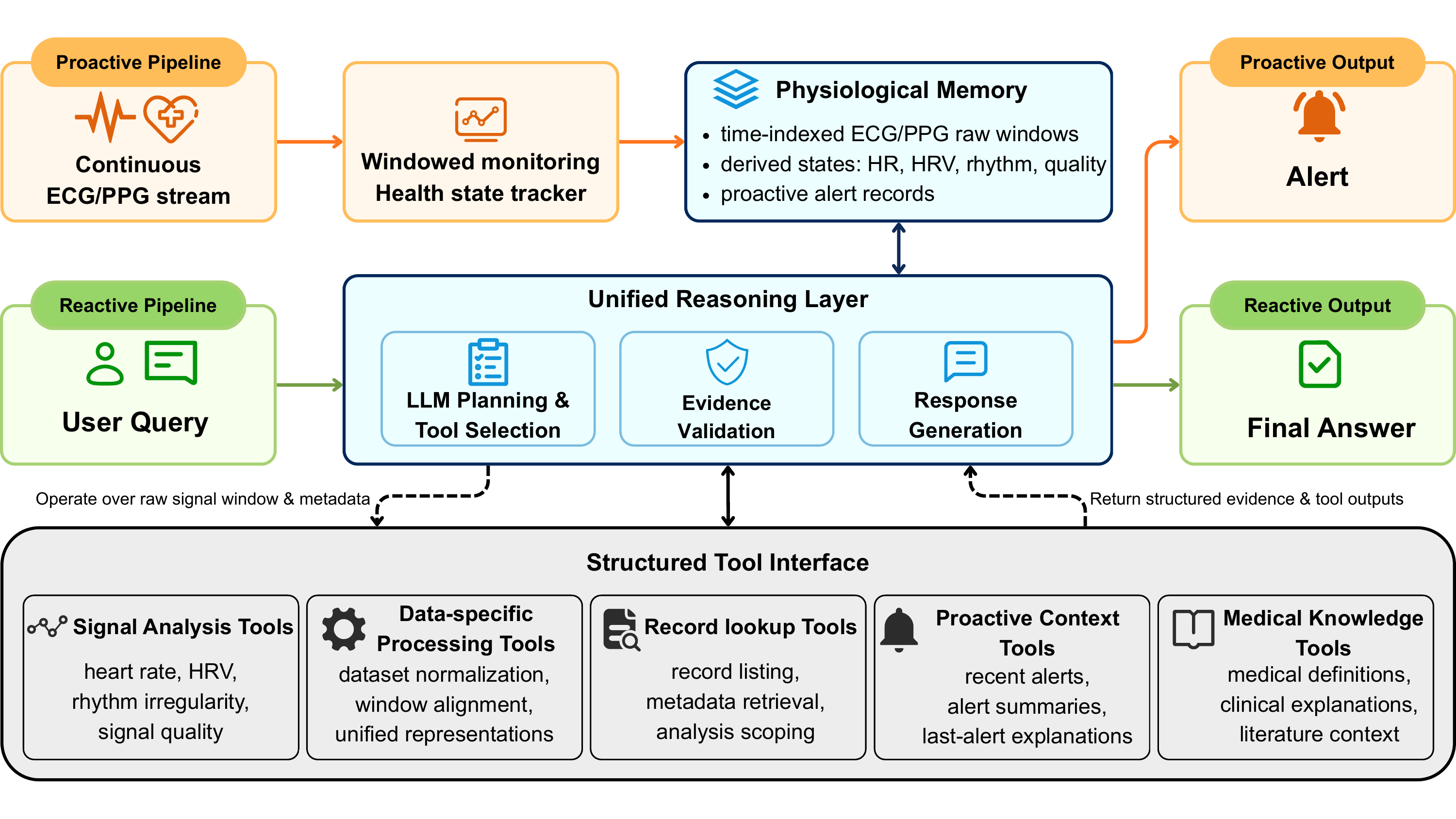} 
    \caption{Overview of \sysname{}, a unified framework for reactive question answering and proactive physiological monitoring via shared planning, tool use, and validation.}
    \label{fig:overview}
\end{figure*}

Building on \dataname{}, we introduce \sysname{}.  As illustrated in Figure~\ref{fig:overview}, \sysname{} supports both reactive question answering and proactive physiological monitoring through a shared memory, reasoning layer and tool registry.

In reactive mode, the system receives a user query \(q\) and uses an LLM-based policy to select tools according to the query and the current physiological memory. These tools retrieve relevant ECG/PPG signals and monitoring context, perform query-specific analysis such as feature extraction or temporal comparison, and return evidence for answer generation. When the collected evidence is incomplete or inconsistent, the system can validate the result and replan before producing the final evidence-grounded answer.

In proactive mode, the system processes non-overlapping 10-second windows sequentially. AF screening is performed using the pretrained single-lead ECG model ECGFounder \citep{ecgfounder}, optionally combined with RR-interval rules, and an alert is triggered when AF is detected in three consecutive windows over 30 seconds. An additional LLM judge can be incorporated to assess the urgency of AF alerts. It is triggered only in the presence of potential anomalies, when the monitoring state changes, including (i) a change in heart-rate range (below 50, 50–100, or above 100 bpm); (ii) a transition between AF and non-AF detections; (iii) AF detection lasting 30 seconds; and (iv) heart rate above 100 bpm for at least 150 seconds within the past five minutes. While anomalies persist without such changes, the judge is triggered again every 200 seconds. The LLM receives structured summaries of the current window and the past five minutes, including the window-level detection results, and can consult clinical guideline sections before making its assessment. The detailed template is provided in Appendix Figure~\ref{fig:proactive-judge-prompt}. The generated alerts are written to memory for reference in subsequent queries. See Appendix~\ref{app:proactive_details} for specific settings.

Thus, the two modes differ mainly in how they are triggered: reactive reasoning is initiated by user questions, while proactive monitoring is initiated by streaming physiological updates. Both modes share the same memory, reasoning layer, and tool set, enabling consistent evidence extraction across question answering and long-term monitoring.

\subsection{System architecture}

As shown in Figure~\ref{fig:overview}, \sysname{} consists of three core components: \emph{a physiological memory, a unified reasoning layer, and a structured tool interface}.

\begin{table}[t]
\centering
\footnotesize
\setlength{\tabcolsep}{4pt}
\caption{Overview of the structured tool registry in \sysname{}, organized into five functional categories and comprising 29 tools in total. Medical knowledge tools query MedlinePlus~\citep{medlineplus_webservice} and PubMed~\citep{ncbi_eutilities, pubmed}. The full registry is provided in Appendix~\ref{app:tool_registry}.}
\begin{tabularx}{\textwidth}{l c c X}
\toprule
Category & Count & Function & Representative tools \\
\midrule
Signal Analysis & 13 & Raw signal feature computation & 
\texttt{analyze\_heart\_rate}, \texttt{ecg\_diagnosis} \\
Data-specific Processing & 3 & Dataset normalization & 
\texttt{analyze\_icentia11k\_ecg\_window\_signal} \\
Record Lookup & 6 & Record listing and metadata access &
\texttt{get\_ecg\_metadata}, \texttt{get\_ppg\_description} \\
Proactive Context & 5 & Proactive alert memory &
\texttt{get\_recent\_alerts}, \texttt{explain\_last\_alert} \\
Medical Knowledge & 2 & External knowledge retrieval &
\texttt{medical\_info\_search}\\
\bottomrule
\end{tabularx}
\label{tab:tool_registry}
\end{table}

\paragraph{Physiological memory.}
The system maintains a compact temporal memory \(\mathcal{M}_t\), defined as \(\mathcal{M}_t = (x_{<t}, s_{<t}, e_{<t})\), where \(x_{<t}\) denotes the history of raw ECG/PPG signal windows, \(s_{<t}\) denotes the streaming health-state tracker updated from previously processed ECG/PPG windows. Specifically, the physiological signals are converted into window-level physiological summaries, including heart rate, beat-to-beat interval features, variability measures, signal quality scores, and rhythm labels that reflect irregularity. \(e_{<t}\) denotes previous proactive alert records, which include the window time, the triggered rule, and an alert justification summary, enabling follow-up questions about recent alerts and their alignment with annotated AF episodes in the evaluation setting. 
In reactive mode, the derived states are computed on demand by raw-signal tools; in proactive mode, they are updated sequentially across windows. 
The memory is continuously updated as new signal windows, derived statistics, and alerts arrive:
\begin{equation}
\mathcal{M}_{t+1} = f_{\text{update}}(\mathcal{M}_t, x_t, s_t, e_t).
\end{equation}

\paragraph{Unified reasoning layer.}
A unified policy interface \(\Pi\) consists of three components: an LLM-based planner for tool selection, an evidence validation module, and an LLM-based response generator. In the reactive setting, the planner selects tools conditioned on the user query \(q\); in the proactive setting, it operates over the evolving physiological memory and monitoring context, and triggers tool invocation when predefined criteria are satisfied.

After tool execution, the system validates the retrieved evidence along four dimensions: plan completeness, tool success, required output fields, and cross-tool consistency. Completeness checks whether all planned steps return results, while success checks whether each tool executes correctly. Field validation ensures that tool-specific outputs, such as \texttt{data}, \texttt{metadata}, \texttt{state}, \texttt{evidence}, or \texttt{explanation}, are present. Consistency validation flags conflicting findings across tools, such as disagreement between heart-rate estimates, rhythm assessments, or signal-quality-dependent outputs.

If validation fails, the system replans by re-prompting the LLM with the original query, previous plan, validation feedback, and allowed tool schema. The final response is generated from the original query, the final tool plan, and the validated tool outputs, including tool names, execution status, and returned structured data or errors. Tool selection is therefore dynamic and driven by the query or physiological state, rather than fixed precomputed signal statistics, enabling on-demand physiological analysis for each interaction. Details of the replanning prompt can be found in Appendix~\ref{app:replanning_prompt}.

\paragraph{Tool interface.}
We define a structured tool registry \(\mathcal{T}\) consisting of five categories and spanning 29 tools in total, summarized in Table~\ref{tab:tool_registry}. All tools return structured outputs (e.g., numerical summaries, temporal trends, or event annotations), enabling explicit computation over raw multimodal physiological signals. The full registry is provided in Appendix~\ref{app:tool_registry}. 

\paragraph{Design principle.}
Unlike existing pipelines that rely on precomputing and storing raw signal statistics~\citep{healthllm, lifeagentbench}, we support dynamic physiological computation conditioned on the user query and the system’s current state, allowing the policy to selectively invoke tools as needed. This design further decouples signal processing from language-based reasoning by externalizing physiological computations (e.g., heart rate estimation and HRV calculation) into dedicated tools, isolating sensor- and device-specific handling within the tool layer so that the policy can focus on evidence integration and temporal reasoning. 

%% file: sections/05_experiments.tex
\paragraph{Backbones.}
\label{sec:baselines}
For all LLM-based experiments, we use DeepSeek-V4 Flash~\citep{deepseekai2026deepseekv4} with temperature 0 as a single, fixed backbone across all methods, isolating the effect of the agent design from backbone capability. It provides strong tool-use and long-context reasoning at low cost. The framework is not tied to this specific model by design, but we do not evaluate other backbones (Appendix~\ref{app:limitations}). We compare \sysname{} with three baselines: i) \emph{Health-LLM}~\citep{healthllm}, which prompts LLMs with natural-language summaries of physiological signals under zero-shot, few-shot, and few-shot CoT settings; ii) \emph{LifeAgent$^*$}\footnote{Code not publicly available; pipeline reproduced.}, adapted from~\citep{lifeagentbench}, a tool-calling agent with deterministic calculation tools; and iii) \emph{PHIA}~\citep{phia}, a ReAct-style health agent that uses code execution as tools.

\paragraph{Evaluations.}
We evaluate context-prompting and agent-style baselines under two context settings: a \emph{leakage-free setting}, where health parameters such as heart rate are derived from signals to simulate real-world scenarios; and an \emph{oracle setting}, where numerical ground-truth (GT) annotations such as heart rate are provided, serving as an upper-bound baseline. All methods receive inputs aligned to the same time range. No model is trained or fine-tuned; each baseline follows the prompting configuration from its original paper, and the context supplied to each tier is detailed in Appendix~\ref{app:monitoringstate_schema}.

For proactive monitoring, we compare four configurations: (i) HR-based rule baseline, which extracts RR intervals from ECG signals to determine whether each window is AF-positive using RR thresholds. (ii) ECGFounder \citep{ecgfounder}, which uses the pretrained single-lead ECG foundation model to classify each window. A window is identified as AF-positive when either the AF score or the AFL score is greater than 0.5. (iii) ECGFounder + RR, which combines the window-level predictions of the RR rules and ECGFounder using an OR rule. A window is identified as AF-positive if either detector predicts AF. If only one detector provides a valid prediction, we use that prediction. (iv) ECGFounder + RR + LLM judge, which extends the third configuration with an LLM-based assessment of AF alert urgency. The LLM is triggered when a rule-based gate identifies potentially abnormal cardiac activity, including an abnormal rhythm classification, an irregular rhythm flag, HR below 50 or above 100 bpm, or a tachycardia ratio of at least 50\% over the preceding 5 minutes. It receives structured physiological evidence, including the fused ECGFounder + RR rhythm classification, the individual RR and ECGFounder outputs, ECGFounder AF/AFL scores, and recent screening evidence. The LLM reassesses the urgency of potential AF events to help determine whether a new alert should be issued when repeated alerts would otherwise be suppressed. Importantly, the LLM does not independently determine AF labels or suppress events that satisfy the deterministic AF screening criteria. Instead, it can adjust alert urgency, thereby influencing alert delivery while leaving AF detection criteria unchanged.

We evaluate five metrics at the window and episode levels. A window is a 10-second segment, while an AF episode is a continuous sequence of windows labeled as AF in the reference annotations. \emph{Window-level AF detection accuracy} is the number of correctly classified AF and non-AF windows divided by the total number of windows with known reference labels. \emph{AF-condition coverage rate} is the number of reference AF episodes containing three consecutive windows that satisfy the AF confirmation criteria, divided by the total number of reference AF episodes. \emph{Alert false positive rate} measures the proportion of reference non-AF windows in which an AF alert is issued; \emph{Alert false negative rate} measures the proportion of reference AF episodes in which no AF alert is issued; \emph{Unmatched AF alerts per hour} measures the number of AF alerts issued during reference non-AF periods per hour of monitoring with known reference labels. Details are reported in Appendix~\ref{app:proactive_details}. The benchmark is split 30\%/70\% into development and test sets\footnote{Code and data: \url{https://github.com/DKSEZD/VitalAgent}}.

%% file: sections/06_results.tex
\begin{table}[t]
\centering
\small
\setlength{\tabcolsep}{3.6pt}
\caption{Accuracy comparison on Tier A and Tier B, comprising 635 and 670 QA pairs, respectively. Under the leakage-free setting, bold marks \sysname{} and the best baseline in each column, and underline indicates the best under the oracle setting.} 
\begin{tabular}{lcccccccc}
\toprule
& \multicolumn{4}{c}{Tier A (635)} & \multicolumn{4}{c}{Tier B (670)} \\
\cmidrule(lr){2-5} \cmidrule(lr){6-9}
Method & Overall & Verify & Choose & Query & Overall & Verify & Choose & Query \\
\midrule
\multicolumn{9}{l}{\emph{Leakage-free setting}} \\
Health-LLM zero-shot & 0.499 & 0.389 & 0.612 & \textbf{0.650} & 0.282 & 0.553 & 0.139 & \textbf{0.095} \\
Health-LLM few-shot & 0.488 & 0.365 & 0.624 & 0.642 & \textbf{0.300} & \textbf{0.561} & 0.191 & \textbf{0.095} \\
Health-LLM few-shot CoT & 0.483 & 0.348 & \textbf{0.635} & \textbf{0.650} & 0.296 & 0.541 & \textbf{0.201} & \textbf{0.095} \\
LifeAgent$^*$ & \textbf{0.545} & \textbf{0.474} & 0.612 & \textbf{0.650} & 0.287 & 0.545 & 0.165 & \textbf{0.095} \\
PHIA & 0.202 & 0.082 & 0.465 & 0.171 & 0.148 & 0.243 & 0.191 & 0.000 \\
\cmidrule(lr){1-9}
\textbf{\sysname{}} (ours) & \textbf{0.860} & \textbf{0.877} & \textbf{0.729} & \textbf{0.992} & \textbf{0.557} & \textbf{0.741} & \textbf{0.443} & \textbf{0.443} \\ \midrule
\multicolumn{9}{l}{\emph{Oracle setting}} \\
Health-LLM zero-shot & 0.784 & 0.678 & 0.841 & \underline{1.000} & 0.588 & 0.773 & 0.232 & \underline{0.688} \\
Health-LLM few-shot & \underline{0.820} & \underline{0.728} & 0.876 & \underline{1.000} & 0.558 & 0.729 & 0.402 & 0.498 \\
Health-LLM few-shot CoT & 0.776 & 0.643 & \underline{0.882} & \underline{1.000} & 0.496 & 0.627 & \underline{0.418} & 0.412 \\
LifeAgent$^*$ & 0.819 & 0.725 & 0.876 & \underline{1.000} & \underline{0.652} & \underline{0.800} & \underline{0.418} & \underline{0.688} \\
PHIA & 0.395 & 0.246 & 0.612 & 0.512 & 0.227 & 0.310 & 0.278 & 0.086 \\
\bottomrule
\end{tabular}
\label{tab:main_results}
\end{table}

\section{Results}
\subsection{Performance in reactive QA}
\noindent\textbf{Baseline comparison.}
For the leakage-free setting in Table~\ref{tab:main_results}, \sysname{} performs best, achieving 0.860 on Tier A and 0.557 on Tier B overall. This improves over the strongest signal-derived baseline by 0.315 on Tier A (LifeAgent$^*$) and 0.257 on Tier B (Health-LLM few-shot) in absolute accuracy. The gap is most evident in Tier B Query, where \sysname{} reaches 0.443, while other baselines all perform poorly.
These results show higher accuracy for the evaluated tool-augmented system than for the tested baselines, but they do not isolate the contribution of LLM-based tool selection from differences in available tools and evidence (see Appendix~\ref{app:limitations} for limitations of this comparison).

The oracle setting with GT states improves the context-prompting baselines, particularly for Health-LLM, with average absolute gains of 0.303 and 0.255 across its three variants for Tiers A and B, respectively. This suggests a strong dependence on the input features. While promising as an upper bound, this setting represents an ideal case and is generally not feasible in real-world applications. \sysname{} shows comparable performance in Tier A and Tier B overall without GT labels, suggesting the value of computing these parameters on demand with tools.

\begin{table}[t]
\centering
\small
\setlength{\tabcolsep}{6pt}
\caption{QA accuracy under injected tool failures on 240 target-balanced Tier-A questions. Validation and replanning yield increasing benefit as the tool-failure rate rises.}
\begin{tabular}{lccc}
\toprule
Tool failure rate & Full & w/o validation & w/o replan \\
\midrule
0\%  & 0.79 & 0.80 & 0.80 \\
15\% & 0.78 & 0.75 & 0.78 \\
30\% & 0.78 & 0.71 & 0.73 \\
50\% & 0.70 & 0.65 & 0.67 \\
\bottomrule
\end{tabular}
\label{tab:ablation_results}
\end{table}

\paragraph{Ablation results. }
To test whether the validation module works properly, we simulate sensor tool-execution failures by replacing selected critical tool outputs with synthetic failures at increasing rates (0/15/30/50\%) using a fixed hash-based injection seed, with the raw physiological signals fixed. On the same target-balanced set of 240 Tier-A questions, we compare the full system against variants without validation and without replanning as shown in Table~\ref{tab:ablation_results}. On clean data, all three variants perform comparably. As the failure rate increases, removing validation or replanning causes noticeably steeper degradation, indicating that both components improve recovery under the evaluated tool-failure setting. When validation reports a failure, it is usually right: 77--85\% of reported failures correspond to genuinely injected tool failures.
Full per-tier ablation results are given in Appendix~\ref{app:phia_analysis},  where replacing the semantic planner with random tool sampling drops Tier-A overall accuracy from 86.0\% to 70.1\% and Tier-B accuracy from 55.7\% to 34.2\%, showing that question-dependent tool selection substantially outperforms random tool selection.

\paragraph{Error analysis. }
\sysname{} fails on 14.0\% of Tier A samples and 44.3\% of Tier B samples. Across both tiers, 75.4\% of failures are \emph{answer mismatches}, where tools execute successfully but the final answer is incorrect, while 24.6\% are \emph{tool selection failures}. Tool selection errors mainly occur in AF- and rhythm-related monitoring-window questions, indicating difficulty in distinguishing window-local analysis from queries requiring longer-range temporal aggregation. This suggests that temporal scope should be more explicitly encoded in both tool descriptions and planning prompts. The largest error category is Tier B WESAD stress questions (122 cases), where the current toolset provides physiological data but lacks a dedicated stress inference module. 
The next most common failures are highest-heart-rate monitoring queries (53 cases), where the system incorrectly aggregates or identifies peak heart rate over time, reflecting challenges in longitudinal numerical reasoning. These results suggest promising directions for future work on more advanced tool design, particularly for complex numerical and temporal reasoning tasks.

\subsection{Proactive performance} 

\begin{table}[t]
\centering
\small
\caption{Proactive monitoring results.
(a) AF detection, condition coverage, and alert performance across four configurations. Detection accuracy and Alert FPR are evaluated over windows, while AF-condition coverage and Alert FNR are evaluated over the 47 reference AF episodes. Unmatched AF alerts/h counts AF alerts during reference non-AF periods per hour of monitoring with known reference labels.
(b) Window-level AF classification. Sensitivity and specificity measure correct classification of reference AF and non-AF windows, respectively; balanced accuracy is their mean. Windows with unknown reference labels are excluded, while unavailable predictions on windows with known labels are counted as incorrect for classification metrics.}
\label{tab:proactive_results}

\textbf{(a) AF detection, condition coverage, and alerts}\\[2pt]
\begingroup
\setlength{\tabcolsep}{2pt}
\renewcommand{\arraystretch}{1.05}
\resizebox{\linewidth}{!}{
\begin{tabular}{@{}lccccc@{}}
\toprule
Configuration
& \begin{tabular}[c]{@{}c@{}}Detection\\accuracy (\%) $\uparrow$\end{tabular}
& \begin{tabular}[c]{@{}c@{}}AF-condition\\coverage (\%) $\uparrow$\end{tabular}
& \begin{tabular}[c]{@{}c@{}}Alert FPR\\(\%) $\downarrow$\end{tabular}
& \begin{tabular}[c]{@{}c@{}}Alert FNR\\(\%) $\downarrow$\end{tabular}
& \begin{tabular}[c]{@{}c@{}}Unmatched\\AF alerts/h $\downarrow$\end{tabular} \\
\midrule
HR-based rule baseline
& 81.61 & 65.96 & 1.216 & \textbf{48.94} & 4.113 \\
ECGFounder
& \textbf{87.37} & 70.21 & \textbf{0.061} & 74.47 & \textbf{0.207} \\
ECGFounder + RR
& 79.12 & \textbf{74.47} & 1.003 & 74.47 & 3.393 \\
ECGFounder + RR + LLM judge
& 79.12 & \textbf{74.47} & 1.144 & 68.09 & 3.869 \\
\bottomrule
\end{tabular}
}
\endgroup

\vspace{6pt}
\textbf{(b) Window-level AF classification}\\[2pt]
\setlength{\tabcolsep}{5pt}
\begin{tabular}{lccc}
\toprule
Detector & Sens. & Spec. & Balanced acc. \\
\midrule
HR-based rule baseline & 78.8\% & 81.8\% & 80.3\% \\
ECGFounder            & 99.8\% & 86.6\% & 93.2\% \\
\bottomrule
\end{tabular}
\label{tab:proactive_results1}
\end{table}

Table~\ref{tab:proactive_results}(a) compares the four configurations across AF detection, condition coverage, and alert performance. ECGFounder alone achieves the highest detection accuracy (87.37\%), the lowest Alert FPR (0.061\%), and the lowest unmatched AF alert rate (0.207/h), but has a substantially higher Alert FNR (74.47\%) than the HR-based rule baseline (48.94\%). Adding RR to ECGFounder improves AF-condition coverage from 70.21\% to 74.47\%. Further adding the LLM judge leaves detection accuracy and condition coverage unchanged and lowers Alert FNR from 74.47\% to 68.09\%, but increases Alert FPR and unmatched AF alerts. Notably, the HR-based baseline achieves the lowest Alert FNR, whereas ECGFounder alone provides the most conservative alerting, with fewer false-positive and unmatched alerts. These results highlight the trade-off between AF detection, condition coverage, and alert delivery, suggesting that combining ECG-based detection with rule-based confirmation can improve coverage, while the LLM judge provides no clear net benefit in this configuration.

To further examine the choice of AF detector, Table~\ref{tab:proactive_results}(b) compares ECGFounder with the HR-based rule baseline. Compared with the HR baseline, ECGFounder increases sensitivity from 78.8\% to 99.8\%, specificity from 81.8\% to 86.6\%, and balanced accuracy from 80.3\% to 93.2\%. These results show that ECGFounder detects more AF windows while correctly identifying a larger proportion of non-AF windows, supporting its use as the AF detector in our framework.

%% file: sections/07_conclusion.tex
This work introduces \sysname{}, a tool-augmented agentic framework for wearable-based mHealth that integrates reactive question answering and proactive physiological monitoring within a shared memory and tool-augmented reasoning architecture, and \dataname{}, which constructs both reactive QA pairs and proactive monitoring data. Extensive experiments demonstrate that \sysname{} consistently outperforms prompt-based and agentic baselines, highlighting the importance of dynamic tool invocation and long-term physiological context for mHealth reasoning. Overall, this work provides a step toward agentic health systems that unify continuous monitoring and interactive reasoning over real-world physiological data.

%% file: sections/08_appendix.tex
\newpage
\appendix
\section*{Appendix}
\section{Dataset sources, licensing, and statistics}
\label{app:dataset_details}

\paragraph{Ethical considerations.}
This work uses publicly available wearable physiological datasets and does not involve the collection of new participant data. We do not attempt to re-identify participants. Nevertheless, physiological signals and derived health information may be sensitive, and appropriate privacy and data-governance practices should be implemented in real-world deployments. By design, raw ECG/PPG waveforms are processed by non-LLM signal-analysis modules and are not included in LLM inputs; the language model receives only structured derived features and contextual summaries. Although this design reduces exposure of raw waveforms, derived health information and patient context may still be sensitive and require appropriate safeguards.

\paragraph{Artifacts, licenses, and intended use.}
This work uses and creates scientific artifacts, including source code, a derived \dataname{}, result logs, public physiological datasets, baseline implementations, software packages, and API-based LLM backbones. The paper cites the creators of the datasets, baselines, software packages, and models used in the experiments. The \dataname{} is derived from four data sources: AF-PPG-ECG, Icentia11k, PPG-DaLiA, and WESAD. We use each artifact consistently with its documented research purpose and access conditions. Icentia11k is released under CC BY-NC-SA 4.0. PPG-DaLiA is released through the UCI Machine Learning Repository under CC BY 4.0. WESAD is an external UCI-listed wearable stress and affect dataset; we follow the linked dataset acknowledgment and licensing terms. The AF-PPG-ECG dataset hosted on Zenodo is listed with an ``Other (Non-Commercial)'' license, so it is used only for non-commercial research benchmarking. We do not redistribute restricted raw signals except as allowed by the source terms, and derived releases should preserve attribution, non-commercial, and research-use constraints from the source artifacts. Accordingly, \dataname{} is released under CC BY-NC-SA 4.0, which preserves the attribution, non-commercial, and share-alike terms of its most restrictive source.

The final \dataname{} contains 1,862 question-answer examples. It covers AF-PPG-ECG, Icentia11k, PPG-DaLiA, and WESAD. The subset is stratified by dataset and difficulty tier: AF-PPG-ECG has 278 Tier-A and 278 Tier-B examples; Icentia11k has 278 Tier-A and 278 Tier-B examples; PPG-DaLiA has 200 Tier-A and 200 Tier-B examples; and WESAD has 150 Tier-A and 200 Tier-B examples. The deterministic development/test split contains 557 development examples and 1,305 test examples. The development split contains 271 Tier-A and 286 Tier-B examples; the test split contains 635 Tier-A and 670 Tier-B examples. The final subset uses selection seed \texttt{42} with a SHA-256 ordering over question identifiers. The development/test split is stratified by dataset and tier and uses SHA-256 ordering with split seed \texttt{dev\_test\_split\_v3}. After development iteration, 33 inspected test items were moved to the development split through a 1:1 strata-preserving swap using SHA-256 ordering with swap seed \texttt{leakage\_repair\_v3}. 

The benchmark covers single-window questions, previous-window comparisons, and monitoring-window aggregates. Target phenomena include atrial-fibrillation presence and burden, heart-rate category and numeric estimates, rhythm changes, stress presence and duration. Because the final full-tier runs evaluate all examples in the tier, the evaluation seed does not change which datapoints are included; it only controls deterministic processing order, and would affect subset selection only when a sample cap or per-target limit is used. 

We release the generated question-answer pairs, the construction and evaluation scripts, and the window indices required to reproduce every item; the underlying raw recordings are not redistributed and must be obtained from the original sources under their respective terms.

\section{Tool registry}
\label{app:tool_registry}
Table~\ref{tab:complete_tool_registry} lists the complete tool registry implemented in our system. Most tools are directly exposed to the agent for signal analysis, window-level data retrieval, temporal comparison, proactive monitoring replay, alert-context access, and external medical knowledge retrieval. The state construction and state access tools are included for completeness, but they are excluded from the agent-facing tool list during evaluation; they are only used offline to construct and query monitoring states when generating the VitalBench benchmark.

\noindent\textit{\underline{Signal analysis tools.}}
We use 13 coding tools that operate on raw ECG/PPG windows and return structured physiological measurements. They implement deterministic computations such as heart rate, HRV, and signal quality, ensuring all low-level features are explicitly derived rather than inferred.

\noindent\textit{\underline{Data-specific processing tools.}}
We use 3 tools that handle heterogeneity across wearable data. They support different modalities (e.g., ECG/PPG), sampling rates, and window lengths (e.g., 30s/60s), while normalizing data-specific formats into unified representations.

\noindent\textit{\underline{Record lookup tools.}}
These tools provide metadata access to available records without performing signal analysis. They support operations such as enumeration of records per modality and retrieval of a specific record's metadata, such as sampling rate, duration, demographic context, together with a human-readable summary. They expose dataset structure to the planner, allowing it to scope subsequent analysis calls.

\noindent\textit{\underline{Proactive context tools.}}
These tools support follow-up QA after proactive monitoring by retrieving stored alert records, summaries, and explanations from prior runs. Unlike record lookup tools, they operate on cached proactive alert outputs rather than raw signal windows or recomputed intervention decisions. This enables explanations of recent alerts and, in evaluation, their alignment with annotated AF transitions. 

\noindent\textit{\underline{Medical knowledge tools.}}
These tools access biomedical knowledge beyond patient data, using MedlinePlus and PubMed for definitions, clinical explanations, and literature context.

\begin{table}[t]
\centering
\small
\setlength{\tabcolsep}{4pt}
\caption{VitalAgent uses 29 agent-facing tools; the full implementation registry contains 41 tools, including VitalBench construction utilities. Categories marked ${\dagger}$ are used for VitalBench construction and are not exposed to the agent at evaluation time.}
\begin{tabularx}{\textwidth}{p{0.25\textwidth} X X}
\toprule
Category & \multicolumn{2}{c}{Tools} \\
\midrule
Signal Analysis
& \texttt{analyze\_heart\_rate} 
& \texttt{analyze\_hrv} \\
& \texttt{analyze\_lead\_morphology} 
& \texttt{analyze\_morphology} \\
& \texttt{ecg\_diagnosis} 
& \texttt{analyze\_pulse\_rate} \\
& \texttt{analyze\_prv} 
& \texttt{assess\_all\_leads\_quality} \\
& \texttt{assess\_ppg\_signal\_quality} \\
& \texttt{assess\_signal\_quality} \\
& \texttt{analyze\_ppg\_rhythm\_irregularity} \\
& \texttt{analyze\_af\_ppg\_ecg\_rhythm\_context} \\
& \texttt{classify\_wesad\_stress\_state} 
& \\

\midrule
Data-specific Processing
& \texttt{analyze\_icentia11k\_ecg\_window\_signal} \\
& \texttt{analyze\_ppg\_dalia\_window\_signal} \\
& \texttt{analyze\_wesad\_window\_signal}
& \\

\midrule
Record Lookup 
& \texttt{get\_ecg\_description} 
& \texttt{get\_ecg\_metadata} \\
& \texttt{get\_ppg\_description}
& \texttt{get\_ppg\_metadata} \\
& \texttt{list\_ecg\_records}
& \texttt{list\_ppg\_records} \\

\midrule
Proactive Context
& \texttt{proactive\_explain\_last\_alert} 
& \texttt{proactive\_load\_patient\_context} \\
& \texttt{proactive\_get\_recent\_alerts} \\
& \texttt{proactive\_list\_patient\_contexts} \\
& \texttt{evaluate\_proactive\_rules} 
& \\

\midrule
Medical Knowledge
& \texttt{medical\_info\_search}
& \texttt{medical\_knowledge} \\

\midrule
State Construction$^{\dagger}$
& \texttt{state\_build\_from\_ecg\_record} \\
& \texttt{state\_build\_from\_ppg\_dalia\_pickle} \\
& \texttt{state\_build\_from\_ppg\_patient} \\
& \texttt{state\_build\_from\_wesad\_pickle} \\

\midrule
State Access$^{\dagger}$
& \texttt{state\_get\_evidence}
& \texttt{state\_get\_longitudinal\_trend} \\
& \texttt{state\_get\_monitoring\_window} 
& \texttt{state\_list\_contexts} \\
& \texttt{state\_load\_monitoring\_states} \\
& \texttt{state\_get\_previous\_monitoring\_state} \\
& \texttt{state\_get\_current\_monitoring\_state} \\
& \texttt{state\_get\_dataset\_capabilities} \\

\midrule
Total & \multicolumn{2}{l}{41 tools (29 agent-facing)} \\
\bottomrule
\end{tabularx}
\label{tab:complete_tool_registry}
\end{table}

\section{Replan prompt}
\label{app:replanning_prompt}

Figure~\ref{fig:replanning-prompt} presents the prompt used when the initial tool plan fails validation. The placeholders are populated with the previous plan, identified validation issues, and the tools available for the corresponding dataset and signal modality. 

The prompt template is shared across operating modes and therefore refers to state access tools; under the benchmark evaluation configuration these tools are excluded from the \texttt{\{tool\_names\}} list supplied at runtime, and the planner is constrained to the listed names, so no state access tool is available to the agent during evaluation.

\begin{figure}[t]
\centering
\begin{minipage}{0.96\textwidth}
\begin{Verbatim}[
  fontsize=\footnotesize,
  breaklines=true,
  breakanywhere=true,
  breaksymbolleft={},
  breaksymbolright={},
  xleftmargin=0.5em,
  xrightmargin=0.5em,
  frame=single,
  framesep=3mm
]
You are the Planner module of a multimodal mHealth Agent. A previous plan
failed validation. Analyse the failure reasons and produce a revised plan.

## Previous plan
{previous_plan}

## Validation issues
{issues}

## Instructions
- Fix the issues identified above.
- You may add, remove, or modify steps.
- Respect the active signal modality implied by the prior plan.
- If the plan is for PPG, do NOT add ECG-only morphology or diagnosis tools.
- If data is genuinely unavailable, include a step that acknowledges the gap.
- If the failed plan used ECG-specific tools for a non-ECG dataset, replace them
  with the relevant state tools.
- If an mHealth-QA WindowLocator is present, use dataset, patient_id,
  window_start_s, and window_end_s as the authoritative tool arguments. For
  state tools, patient_id may be used as subject_id.
- If dataset support is unclear, add `state_get_dataset_capabilities` before
  answering target-specific questions.
- Do not infer AF, rhythm, stress, or activity when the active dataset
  capabilities say the target is unsupported.

## Allowed tool names (STRICT)
You MUST use tool names from this exact list only:
{tool_names}

If a needed tool is unavailable, do not invent a new tool name. Instead, use the
closest available tool(s) from the allowed list.

{tools_description}

Return ONLY valid JSON with the same schema as before (no markdown fences).
\end{Verbatim}
\end{minipage}
\caption{Prompt used for replanning after validation failure.}
\label{fig:replanning-prompt}
\end{figure}

\section{Proactive monitoring details}
\label{app:proactive_details}

\textbf{AF screening.}
The HR-based rule baseline marks a window as AF-positive when the RR-interval rules (Appendix~\ref{app:implementation}) classify it as AF. For ECGFounder, we apply the pretrained single-lead ECGFounder model~\citep{ecgfounder} to each complete 10-second ECG window and mark a window as AF-positive when either its AF or AFL score exceeds 0.5. ECGFounder + RR combines the two window-level results with the OR rule described in Section~\ref{sec:baselines}. AF is confirmed after three consecutive windows (30 seconds) that each provide confirmation evidence: an AF classification by the RR rules in the HR-based rule baseline, the higher of the two ECGFounder scores at or above 0.9 in ECGFounder, and either of the two in ECGFounder + RR, so that consecutive windows may be supported by different detectors.

After confirmation, the AF condition is active whenever the current window is AF-positive. A valid non-positive window makes the condition inactive, but the confirmed state is retained until non-positive results continue for 20 seconds. Unavailable input or a time gap clears both the confirmed state and the accumulated evidence. Neither is treated as evidence of recovery.

The ECGFounder thresholds are engineering settings selected on independent development data, and the RR-interval thresholds were calibrated as described in Appendix~\ref{app:implementation}; neither is a calibrated clinical probability. The confirmation and recovery durations are the same for all four configurations.

\textbf{LLM judge.}
The LLM judge is used only in ECGFounder + RR + LLM judge. A rule-based gate selects windows for further assessment using the available heart-rate or rhythm evidence. A window passes the gate if it has an abnormal rhythm class, an irregular rhythm flag, HR below 50 or above 100 bpm, or a tachycardia ratio of at least 50\% in the trailing 5-minute window. No LLM request is made when both heart-rate and rhythm evidence are unavailable.

Passing the gate does not necessarily trigger a new LLM request. A valid cached decision is reused while the relevant input categories remain unchanged. These categories include rhythm class, rhythm regularity, heart-rate band, whether consecutive AF-positive results have lasted 30 seconds, whether the tachycardia ratio has reached 50\%, and evidence availability. The system reassesses the decision when these categories change or after 20 windows of cache reuse. A failed request or invalid response is retried once. If the retry also fails, the result is not cached as a valid decision, and further requests are paused for 20 windows.

The judge receives structured physiological evidence, including evidence availability, the fused AF/NonAF rhythm class of ECGFounder + RR, the separate RR and ECGFounder results, and the ECGFounder AF/AFL scores. Reference annotations are used only for evaluation and are excluded from model inputs. The input states that these scores are neither diagnoses nor calibrated confidence estimates. AF events proposed by the judge must satisfy the same screening condition described above. If the AF condition is inactive, the proposed AF event is removed. If it is active, the judge may increase its urgency but cannot suppress the event generated by deterministic screening. Figure~\ref{fig:proactive-judge-prompt} shows the judge's system prompt, which is shared with other monitoring settings and therefore also refers to heart-rate rules and rhythm classes that are not used in these experiments.

\textbf{Alerts.}
The system keeps an alert record for the AF event. It sends an alert when the event is first registered or when its urgency exceeds the highest level already reported for that event; since deterministic screening always assigns low urgency, such urgency upgrades occur only with the LLM judge. Otherwise, repeated alerts are suppressed.

In all four configurations, the system closes the AF alert record only after the AF condition has been absent for 20 consecutive seconds with available NonAF detection results. When the LLM judge is enabled, its decision must also be valid in each recovery window. Missing input is not evidence of recovery. Once the record is closed, a later occurrence can generate a new alert.

\textbf{Episode breakdown.}
All detections, confirmations, and alerts are timed at the end of the 10-second window in which they occur. Under ECGFounder, far more episodes satisfy the AF condition (33/47) than receive an alert (12/47). All 47 reference episodes contained AF-positive detections and fall into three groups: 14 did not contain three consecutive windows satisfying the AF confirmation criteria; 21 satisfied these criteria but had no new alert because an earlier AF alert record in the same recording remained open; and 12 contained a new AF alert. Therefore, the absence of a new alert does not necessarily indicate a missed detection.

The 11 episodes shorter than 30 seconds cannot satisfy the AF condition; among the 36 episodes lasting at least 30 seconds, 33 satisfied it within the interval. For AFIB, detection coverage, AF-condition coverage, and new AF alert coverage were 40/40, 27/40, and 8/40, respectively. For AFL, the corresponding counts were 7/7, 6/7, and 4/7.

ECGFounder + RR produced identical detections, confirmations, and new alerts with and without the LLM judge. The judge raised AF alert coverage from 8/40 to 10/40 for AFIB and from 4/7 to 5/7 for AFL through urgency upgrades, and for the 12 episodes alerted in both configurations, the first AF alert times were identical. AF alerts in windows with unknown reference labels are not counted in Alert FPR or unmatched AF alerts/h; there were 41 such alerts under ECGFounder + RR and 53 under ECGFounder + RR + LLM judge.

\textbf{Supplementary results.} Per-cohort window-level results for Table~\ref{tab:proactive_results}(b) are given in Table~\ref{tab:proactive_window_cohort}.

\begin{table}[h]
\centering
\small
\caption{Per-cohort window-level AF classification, using the same definitions as Table~\ref{tab:proactive_results}(b). Windows with unknown reference labels (2,980, all from Icentia11k) are excluded.}
\label{tab:proactive_window_cohort}
\setlength{\tabcolsep}{4pt}
\begin{tabular}{lcccccc}
\toprule
& \multicolumn{3}{c}{HR-based rule baseline} & \multicolumn{3}{c}{ECGFounder} \\
\cmidrule(lr){2-4}\cmidrule(lr){5-7}
Cohort & Sens. & Spec. & Bal. acc. & Sens. & Spec. & Bal. acc. \\
\midrule
Icentia11k       & 71.9\% & 85.7\% & 78.8\% & 100.0\% & 89.6\% & 94.8\% \\
AF-PPG-ECG (ECG) & 84.3\% & 80.5\% & 82.4\% & 99.7\%  & 85.6\% & 92.7\% \\
\midrule
Total            & 78.8\% & 81.8\% & 80.3\% & 99.8\%  & 86.6\% & 93.2\% \\
\bottomrule
\end{tabular}
\end{table}

\begin{figure}[t]
\centering
\begin{minipage}{0.98\textwidth}
\begin{Verbatim}[
  fontsize=\scriptsize,
  breaklines=true,
  breakanywhere=true,
  breaksymbolleft={},
  breaksymbolright={},
  xleftmargin=0.1em,
  xrightmargin=0.1em,
  frame=single,
  framesep=3mm
]
On every window of {signal_phrase} data, the agent evaluates whether to alert the user. Two layers make that call: a guideline-grounded rule layer (handles unambiguous extremes and sustained tachycardia at rest) and you, the LLM judge. Your role is to handle the events where the guidelines' recommendations are conditional on context the rule layer cannot encode.

## Your input

On each invocation you receive a structured snapshot of the user's current health state:
- `hr_bpm`: current heart rate, or null
- `rhythm_class`: "N", "AF", "AFL", "Other", or null
- `af_episode_duration_s`: seconds the current AF episode has lasted, or null if not in AF
- `tachycardia_ratio_5min`: fraction of the trailing 5-minute window with HR > 100 bpm (0.0 to 1.0), or null
- `tachycardia_sample_count`: number of HR samples in that trailing window

You also receive a small patient context (known conditions, if any), which may be empty in Phase 1.

## Your tools

You have one tool: `read_guideline_section(guideline_id, section_id)`. Use it when you need the full text of a specific guideline section. The executive summaries of all guidelines are already in your context below; fetch full text only when a decision genuinely depends on detail the summary does not cover.

Limit yourself to at most 3 tool calls per decision. Each unnecessary call costs latency and tokens.

## Your output

Respond with a single JSON object (no markdown fences, no prose outside the JSON) of this shape:

{{
  "intervene": true | false,
  "urgency": "none" | "low" | "medium" | "high" | "critical",
  "event_types": ["af"],
  "reason": "one-sentence clinical justification",
  "advice": "short plain-language guidance for the user",
  "cited_sections": [
    {{"guideline_id": "af-2023", "section_id": "ahre-5min-to-24h"}}
  ]
}}

Rules for each field:

- `intervene`: true only if the state warrants user-facing alerting. When false, the other fields should be empty strings or empty lists.
- `event_types`: list all distinct notification-worthy conditions. When intervening, include at least one of "af", "bradycardia", "tachycardia", "irregular_rhythm", or "other"; otherwise return []. Use "af" for an AF-related alert, not a definitive diagnosis. These stable causes are independent of which guideline sections you cite.
- `urgency`: pick the lowest tier consistent with the guidance. "low" is informational, "medium" is "consider seeing a clinician soon", "high" is "see a clinician promptly", "critical" is "seek immediate medical attention". Prefer "medium" unless the evidence genuinely warrants escalation.
- `reason`: a single sentence naming the clinical concern in plain terms (e.g. "Atrial fibrillation episode of 18 minutes in a user without a known AF history, within the guideline's AHRE decision band.").
- `advice`: short, non-diagnostic user-facing guidance. Never recommend medication changes. Never give a diagnosis. Always route the user to a qualified clinician when the recommended action is more than reassurance.
- `cited_sections`: zero or more guideline sections your reasoning relied on. Use the exact guideline_id and section_id from the summaries below. Empty list is acceptable for straightforward cases.

## Consumer-device constraints

This system is a consumer monitoring agent, not a diagnostic device. Per FDA consumer-device framework:
- Never provide a diagnosis.
- Never recommend starting, stopping, or changing medication.
- When advice is needed beyond reassurance, the action ceiling is "consult a healthcare professional".

## When to return no alert

If the signals are within normal ranges, or within a range that is expected for routine activity, or if the evidence is insufficient to justify concern, return `intervene: false` with empty fields. Alert fatigue is a real harm; do not alert on every marginal reading.

## Clinical guidelines (executive summaries)

{guideline_summaries}
\end{Verbatim}
\end{minipage}
\caption{Prompt used for proactive alert judgment.}
\label{fig:proactive-judge-prompt}
\end{figure}

\begin{table*}[t]
\centering
\small
\setlength{\tabcolsep}{6pt}
\caption{Ablation results of \sysname{}. “w/o planner” replaces the semantic planner with random tool sampling; “w/o validation” removes intermediate output checking; and “w/o replan” disables tool revision. V/C/Q denote verify, choose, and query. Bold and underline mark the best and second-best results in each column.}
\vspace{4pt}
\begin{tabular}{lrrrrrrrr}
\toprule
& \multicolumn{4}{c}{Tier A} & \multicolumn{4}{c}{Tier B} \\
\cmidrule(lr){2-5} \cmidrule(lr){6-9}
Variant & Overall & V & C & Q & Overall & V & C & Q \\
\midrule
\sysname{} & \textbf{0.860} & 0.877 & \textbf{0.729} & \underline{0.992} & 0.557 & 0.741 & 0.443 & \textbf{0.443} \\
w/o planner & 0.701 & \textbf{0.892} & 0.576 & 0.341 & 0.342 & 0.537 & 0.376 & 0.086 \\
w/o validation & 0.849 & 0.860 & \underline{0.718} & \textbf{1.000} & \textbf{0.569} & \textbf{0.761} & \underline{0.459} & \textbf{0.443} \\
w/o replan & \underline{0.855} & \underline{0.880} & 0.706 & \underline{0.992} & \underline{0.563} & \underline{0.757} & \textbf{0.469} & \underline{0.421} \\
\bottomrule
\end{tabular}
\label{tab:ablation_full}
\end{table*}

\section{MonitoringState schema and leakage control}
\label{app:monitoringstate_schema}
Tier-A baselines receive current- and previous-window estimates, and Tier-B baselines receive aggregated estimates over the full monitoring window; both \sysname{} and all baselines are given the same time range.

Table~\ref{tab:monitoring_state_visible} summarizes the main \texttt{MonitoringState} fields that are visible under the benchmark evaluation constraints. Table~\ref{tab:monitoring_state_hidden} summarizes annotation- or label-derived fields used for ground-truth derivation and audit; these fields are hidden from evaluated methods during fair evaluation.

\begin{table}[t]
\centering
\small
\setlength{\tabcolsep}{3pt}
\caption{\texttt{MonitoringState} schema fields and benchmark visibility handling. Under the oracle setting, baselines see leakage-filtered state records with annotation labels removed and numerical state fields retained. Under the leakage-free setting, target-equivalent numerical state fields are replaced by leakage-safe signal-derived computations. VitalAgent uses sanitized raw-signal tools rather than direct \texttt{MonitoringState} JSONL access.}
\begin{tabularx}{\textwidth}{>{\raggedright\arraybackslash\hyphenpenalty=10000\exhyphenpenalty=10000}p{0.14\textwidth} l l >{\raggedright\arraybackslash}X}
\toprule
Category & Field & Type & Description \\
\midrule
Identity & \texttt{state\_id} & string & Unique identifier for the monitoring state. \\
 & \texttt{patient\_id} & string & Patient or subject identifier. \\
 & \texttt{subject\_id} & string/null & Dataset subject identifier. \\
 & \texttt{recording\_id} & string/null & Recording/session identifier. \\
 & \texttt{dataset} & string & Source dataset name. \\
 & \texttt{modality} & string & Primary modality represented by the state. \\
 & \texttt{window\_index} & integer/null & Index of the monitoring window in the patient timeline. \\
 & \texttt{window\_start\_s} & float/null & Window start time in seconds. \\
 & \texttt{window\_end\_s} & float/null & Window end time in seconds. \\
 & \texttt{window\_duration\_s} & float/null & Window duration in seconds. \\
 & \texttt{recording\_duration\_s} & float/null & Total recording duration in seconds when available. \\
\midrule
 Signal-derived & \texttt{hr\_bpm} & float/null & Heart rate for the current window. \\
 cardiac features & \texttt{mean\_hr\_bpm} & float/null & Mean heart rate over the window or interval. \\
 & \texttt{max\_hr\_bpm} & float/null & Maximum heart rate over the window or interval. \\
 & \texttt{previous\_hr\_bpm} & float/null & Heart rate from the previous comparable window. \\
 & \texttt{baseline\_resting\_hr} & float/null & Subject-specific resting heart-rate baseline when available. \\
 & \texttt{hr\_deviation\_from\_baseline} & float/null & Deviation of current heart rate from the resting baseline. \\
 & \texttt{sdnn\_ms} & float/null & SDNN heart-rate-variability or pulse-rate-variability statistic. \\
 & \texttt{rmssd\_ms} & float/null & RMSSD heart-rate-variability or pulse-rate-variability statistic. \\
 & \texttt{signal\_quality\_score} & float/null & Signal quality score when computed by the corresponding processor. \\
 & \texttt{motion\_level} & string/null & Coarse motion level for PPG or wearable windows when available. \\
\midrule
Short-term aggregate & \texttt{mean\_hr\_5min} & float/null & Mean heart rate over the short-term monitoring scale. \\
& \texttt{tachycardia\_ratio\_5min} & float/null & Fraction of short-term windows meeting tachycardia criteria. \\
\midrule
Proactive alert context & \texttt{alert\_triggered} & boolean & Whether the proactive pipeline emitted an alert for the state. \\
& \texttt{alert\_rule} & string/null & Identifier of the triggered proactive rule. \\
& \texttt{alert\_reason} & string/null & Evidence summary for the alert. \\
& \texttt{urgency} & string/null & Rule-defined urgency level. \\
\midrule
Auxiliary containers & \texttt{metadata} & object & Dataset-level or processing metadata. Leakage-prone keys are filtered during fair evaluation. \\
& \texttt{dataset\_specific} & object & Dataset-specific fields not shared across all sources. Leakage-prone keys are filtered during fair evaluation. \\
\bottomrule
\end{tabularx}
\label{tab:monitoring_state_visible}
\end{table}

\begin{table*}[t]
\centering
\small
\setlength{\tabcolsep}{1mm}
\caption{\texttt{MonitoringState} fields hidden from evaluated methods during fair evaluation. These fields are used only for ground-truth derivation and audit; they are excluded from agent tool outputs and from baseline prompts.}
\vspace{4pt}
\begin{tabularx}{\textwidth}{>{\raggedright\arraybackslash\hyphenpenalty=10000\exhyphenpenalty=10000}p{0.18\textwidth} l l >{\raggedright\arraybackslash}X}
\toprule
Category & Field & Type & Description \\
\midrule
Annotation-derived current state & \texttt{rhythm\_class} & string/null & Reference rhythm class aligned to the current window. \\
& \texttt{stress\_label} & string/null & Reference stress or affective state. \\
& \texttt{protocol\_label\_id} & integer/null & Dataset protocol label identifier. \\
& \texttt{activity\_label} & string/null & Activity label when provided by the dataset. \\
& \texttt{ecg\_reference\_hr\_bpm} & float/null & ECG-derived reference heart rate used as ground truth for PPG-DaLiA pulse-rate questions. \\
\midrule
Previous-window annotation context & \texttt{previous\_rhythm\_class} & string/null & Previous-window rhythm class. \\
& \texttt{previous\_stress\_label} & string/null & Previous-window stress label. \\
& \texttt{previous\_protocol\_label\_id} & integer/null & Previous-window protocol label identifier. \\
& \texttt{previous\_activity\_label} & string/null & Previous-window activity label. \\
\midrule
Annotation-derived aggregates & \texttt{af\_burden\_ratio} & float/null & Fraction of interval covered by AF reference annotations. \\
& \texttt{af\_episode\_duration\_s} & float/null & AF episode duration in seconds when available. \\
& \texttt{rhythm\_transition\_count} & integer/null & Number of rhythm transitions over the monitoring interval. \\
& \texttt{rhythm\_transition\_count\_per\_hour} & float/null & Rhythm transition frequency per hour. \\
& \texttt{stress\_burden\_ratio} & float/null & Fraction of interval labeled as stress. \\
& \texttt{stress\_duration\_s} & float/null & Duration labeled as stress in seconds. \\
& \texttt{dominant\_stress\_label} & string/null & Dominant stress label over the monitoring interval. \\
& \texttt{stress\_transition\_count} & integer/null & Number of stress-label transitions. \\
& \texttt{stress\_transition\_count\_per\_hour} & float/null & Stress-label transition frequency per hour. \\
\bottomrule
\end{tabularx}
\label{tab:monitoring_state_hidden}
\end{table*}

\section{Additional ablation and baseline analysis}
\label{app:phia_analysis}

\paragraph{Full ablation results.}
Table~\ref{tab:ablation_full} reports ablations over the main components of the tool-usage pipeline: planning, validation, and replanning. Replacing the planner with random tool sampling leads to substantial accuracy drops, from 86.0\% to 70.1\% on Tier A and from 55.7\% to 34.2\% on Tier B. The largest decline occurs in Tier A Query, where accuracy decreases from 99.2\% to 34.1\%, showing that question-dependent tool selection is central to the agent's performance. This ablation compares the LLM planner only with random tool sampling, not with a deterministic task-aware router (Appendix~\ref{app:limitations}).

By contrast, removing validation or replanning changes overall accuracy by less than 1.3\%. This is partly because replanning is rarely needed: only 0.4\% of samples require a second tool plan. Validation and replanning therefore act as safeguards for harder cases such as missing data, tool failures, or ambiguous queries, which is what the injected-failure experiment in the main text evaluates.

\paragraph{Additional analysis of PHIA.}
PHIA performs less competitively than the other baselines in our setting. We attribute this mainly to a mismatch between its original design assumptions and the evaluation setting of \dataname{}. PHIA is designed for personal health reasoning over richer wearable records, where inputs often include higher-level behavioral or lifestyle variables such as steps, activity summaries, and calories. In contrast, \dataname{} focuses on short-window physiological signal interpretation and longitudinal aggregation over ECG/PPG-derived estimates.

As a result, PHIA receives a less natural input format in our benchmark. The leakage-free setting provides compact physiological estimates rather than the richer personal health records used in PHIA's original setting. This limits PHIA's ability to exploit its intended reasoning structure. Therefore, its lower performance should not be interpreted as a general weakness of PHIA, but rather as evidence that systems designed for personal health records do not directly transfer to raw-signal-centered physiological QA without additional signal-analysis tools.

\section{Implementation details and reproducibility}
\label{app:implementation}

\paragraph{Computational experiments and budget.}
The computational experiments evaluate \sysname{} against direct LLM and agent-style baselines on \dataname{}, and evaluate a proactive AF detection pipeline on retrospective ECG streams. Unless otherwise specified, \dataname{} final test experiments use the held-out test split once per condition and evaluate the full tier: 635 Tier-A examples or 670 Tier-B examples. The checked final test runs set \texttt{miss-ratio=0.25}, but this parameter is inert under the raw data mode used for all reported experiments, in which the agent operates directly on raw signal windows; it applies only to state-backed modes. Development pilot runs used smaller capped subsets for iteration. The checked local configuration sets the shared LLM backbone to \texttt{deepseek-v4-flash}, and the LLM temperature to 0. No \texttt{LLM\_MAX\_TOKENS} override was found in the checked local environment; the code default is 2048 output tokens unless overridden by an experiment command or environment variable.

All experiments were run as API-inference and local signal-processing jobs. We did not train or fine-tune any neural model. The proactive monitoring uses pretrained ECGFounder weights (SHA-256 prefix f863a388), with inference performed exclusively on the CPU; no training or fine-tuning was performed. For hosted proprietary API models, exact parameter counts and server-side GPU hours are not publicly disclosed. Experiments were run on two machines. Machine~A (macOS~26.2, build~25C56; Apple~M4, 10~cores; 16\,GB RAM) was used for the proactive monitoring experiments and the ablation and robustness experiments. Machine~B (Windows~10~Pro 22H2, build~19045.6466; AMD~Ryzen~9~5950X, 16~cores / 32~threads; 64\,GB RAM) was used for the main reactive question-answering experiments. Both used CPU-only execution for non-API processing; no CUDA device was available in either environment. We therefore do not claim a local GPU training budget or a complete provider-side compute budget. The full-tier planner, validation, and replanning ablations were run on Machine~A, whereas the main results were produced on Machine~B; the two machines differ only in the platform used for local signal processing, since all LLM inference is served by the same API endpoint at temperature 0.

\paragraph{Experimental settings and software.}
The main experimental settings are fixed evaluation and inference settings rather than learned hyperparameters: dataset split, tier, condition, maximum sample count, missing-context ratio, evaluation run seed, agent data mode, LLM model, temperature, and maximum output tokens. These settings are fixed before reporting test results and are documented in the evaluation commands and result metadata. For full-tier test runs, the evaluation run seed is retained for reproducible ordering but does not alter the datapoint set. Signal-processing behavior is implemented in the released tool code, including any task-specific thresholds and package calls. The checked Python environment uses Python 3.10.11 with OpenAI client 2.26.0, NumPy 2.2.6, pandas 2.3.3, scikit-learn 1.7.2, wfdb 4.3.1, NeuroKit2 0.2.13, h5py 3.16.0, PyTorch 2.11.0, pydantic 2.12.5, and openpyxl 3.1.5.

\paragraph{Threshold calibration.}
The RR-interval AF classifier, used by the reactive signal-analysis tools and, in proactive monitoring, by the HR-based rule baseline and both ECGFounder + RR configurations, labels a window as AF when coefficient of variation exceeds 0.06, the normalized Shannon entropy of $\Delta$RR exceeds 0.50, and the turning-point ratio lies between 0.45 and 0.80; a window with coefficient of variation below 0.05 is labeled normal sinus rhythm, and the rest are labeled \texttt{Other}. These thresholds were calibrated once on a single Icentia11k segment (patient 01182, segment 00) and then frozen. A different segment of the same patient (segment 01) is part of the Icentia11k evaluation cohort, so calibration and evaluation are not fully participant-disjoint for that cohort.

\paragraph{Proactive monitoring setup and AF screening selection.} 
Each recording is segmented using processed, non-overlapping 10-second windows, resulting in 419 windows for each recording in Icentia11k and a maximum of 500 windows for each recording in AF-PPG-ECG. The recordings in AF-PPG-ECG were cut off at 500 windows to limit the data scale for processing. 
There are 32,475 windows in total. The entry condition of the AF screening was selected on development data outside the evaluation cohort: the first 30 minutes of 12 records from the MIT-BIH Atrial Fibrillation Database~\citep{moody1983new, pollard2026physionet} and a predefined panel of boundary and noise cases. We compared a small set of predefined candidates, selected the simplest high-score entry rule, and froze it before confirmation; the evaluation cohort was used only to diagnose error sources of an earlier screening variant, not to select thresholds. A single confirmation on 11 further records from the same database (5.5 hours) retained AF-condition coverage of the longer episodes (5/5) but did not meet the predefined criterion of producing no more episode-unmatched alerts than a reference configuration combining RR-interval rhythm classification with heart-rate rules (3 vs. 1 alerts of any type). The final experiments nevertheless use this frozen configuration, so its episode-unmatched alert behavior has not been independently confirmed.

\paragraph{Use of the development split.}
The development split was used for iterative system development rather than continuous hyperparameter optimization: constraining answers to the allowed label set, tightening evidence requirements for AF and rhythm questions, standardizing numeric output so the evaluator reads a single leading value, correcting the previous-window heart-rate comparison to a 5\,bpm meaningful-change criterion, aligning analysis windows with benchmark windows, fixing previous-window routing for Icentia11k, adjusting planner and tool routing, and extending leakage filtering. Baseline inference settings were also adjusted to avoid handicapping the baselines; for example, the maximum output length for LifeAgent$^*$ was increased from 32 to 512 tokens. The test split was used only for the final reported runs. The released archive contains the \sysname{} implementation, the \dataname{} construction and evaluation scripts, and the benchmark itself. Baseline implementations are not redistributed: Health-LLM and PHIA are available from
their original releases, and our LifeAgent$^*$ pipeline was reproduced from the paper description, as the original code is not public.

\paragraph{Ablation and robustness settings.} 
The tool-failure injection experiments (Table~\ref{tab:ablation_results}), including their clean-data condition, and the tool-invocation comparisons are run on a target-balanced subset of 240 Tier-A questions; the planning, validation, and replanning ablations in Table~\ref{tab:ablation_full} use the full test tiers. The subset is drawn from the 635 Tier-A questions in the held-out test split by target-stratified balanced random sampling without replacement: candidates are shuffled with random seed \texttt{42} and capped at 30 questions per evaluation target. Tier A spans eight targets, each with at least 30 available questions, giving $8 \times 30 = 240$ questions. All conditions run on this subset, including all tool-failure injection rates, reuse the same 240 question identifiers, so comparisons across conditions are paired. The no-planning ablation additionally sets a maximum immediate tool-action budget. Tool failures are injected at rates of 0, 15, 30, and 50\% at the evaluator level. For each critical tool call, a deterministic pseudo-random value is derived from seed \texttt{42}, the question identifier, the tool name, and the call index; when it falls below the injection rate, the tool executes normally but its result is replaced by a structured \texttt{success=False} failure before reaching the validation layer. Raw signals and ground-truth annotations are never modified. Note that this is distinct from the missing-context ratio discussed above, which is inert under raw data mode. The full system detects the failure at the validation gate and performs at most one replanning attempt to reach an alternative tool or evidence path; the \texttt{no-validation} condition does not detect the failure, and the \texttt{no-replan} condition detects it but cannot attempt recovery. Injection decisions are reproducible for identical calls across runs, but different conditions may produce different plans, so comparisons are paired at the question level rather than at the level of individual tool calls.

\section{Limitations and responsible research statement}
\label{app:limitations}
\paragraph{Limitations.}
We acknowledge several limitations. First, \sysname{} is evaluated primarily on ECG and PPG. Although these modalities are widely available in wearable devices and cover cardiovascular and stress-related applications, extending the framework to additional signals, such as respiration and skin temperature, may support a broader range of monitoring tasks.

Second, the proactive component relies on predefined, domain-informed criteria to initiate downstream alert reasoning. This design provides transparency and controllability, but does not adapt alerting policies to individual physiological patterns or risk profiles. Developing personalized policies while preserving safety and interpretability remains an important direction.

Third, our deployment analysis is limited to feasibility estimates rather than systematic profiling. The deterministic rule layer and ECGFounder inference run on each window, whereas an actual LLM call occurs only after a cache refresh and when a non-LLM coarse gate identifies a potentially abnormal state. LLM inference cost is therefore not incurred for every monitoring window, but may still increase with monitoring duration and event frequency. Raw-sample storage also grows linearly: 24 hours of single-channel 250\,Hz ECG requires approximately 43\,MB at 16-bit precision, or 86\,MB at float32 precision, excluding metadata and derived states. End-to-end latency, memory use, and on-device optimization remain to be evaluated.

Fourth, the reactive questions are written from a fixed set of templates, each with one canonical question and two paraphrases (Section~\ref{sec:reactiveQA}). These questions are used only to build the benchmark; no model is trained or fine-tuned on them. However, the development and test splits use the same templates and wordings, and we used the development split to adjust tool routing and answer rules (Appendix~\ref{app:implementation}). As a result, the test set does not show how well a system handles questions worded in new ways. The fixed question set makes evaluation reproducible, but it may favor systems tuned to its structure, so performance on \dataname{} does not show that a system can handle free-form questions from real users. In addition, several Tier-B questions are answered from the whole recording (AF-PPG-ECG) or the whole labeled study protocol (WESAD) rather than from the time window being asked about. Their answers therefore vary little within each dataset, and accuracy on these questions mainly reflects whether a system agrees with these largely fixed answers.

Fifth, our comparisons cannot fully separate the sources of performance. All LLM-based methods use the same single backbone, so that differences between methods do not come from using different LLMs; however, this does not show whether the results hold for other LLMs. We also do not compare against a conventional pipeline that uses fixed rules, rather than an LLM, to interpret questions and choose the same tools. Because the questions follow fixed templates shared across splits, such a rule-based pipeline could map each test question to its target and might perform comparably on this benchmark. Our results therefore do not separate the contribution of the LLM from that of the specialized tools, and do not show that an LLM is necessary for these tasks; answering this requires evaluation with free-form user questions, which we leave to future work. Similarly, the robustness experiment (Table~\ref{tab:ablation_results}) simulates failures in tool outputs while keeping the raw signals unchanged; it does not test robustness to noisy or missing signal segments.

Finally, although \dataname{} contains long-duration recordings for both reactive and proactive evaluation, it remains a retrospective, controlled benchmark. We do not conduct prospective clinical validation, clinician or user studies, or evaluate downstream patient outcomes. Real-world deployment would additionally involve noisier signals, missing data, device heterogeneity, population shift, and user-specific variability. Future work should evaluate these factors together with false-alert burden, missed events, and appropriate human oversight.

\paragraph{Potential risks.}
\sysname{} and the associated \dataname{} are research artifacts for retrospective evaluation of mobile-health monitoring agents. They are not intended for diagnosis, triage, treatment decisions, emergency response, or replacement of clinical care. Potential risks include incorrect or over-confident answers to health questions, false reassurance from missed abnormal events, user anxiety or alarm fatigue from false alerts, privacy risks from longitudinal physiological signals, and limited generalization across populations, sensors, recording conditions, and clinical settings. These risks are especially important because ECG, PPG, and other wearable physiological signals may be interpreted by users as medically actionable.

We reduce these risks in the study design by using retrospective public or externally released research datasets, separating benchmark answers from clinical recommendations, reporting false-alert behavior and failure modes, and treating proactive monitoring as an alerting experiment rather than a medical device claim. The agent is evaluated with constrained tools and fixed answer formats. Where possible, annotation-like fields and hidden reference labels are excluded from model-facing inputs to reduce leakage; dataset-local identifiers are used only as technical locators. Any real-world deployment would require prospective clinical validation, regulatory and privacy review, user-facing limitations, and human oversight.

\paragraph{Privacy and human subjects.}
We did not recruit participants, collect new human-subject data, run a user study, or hire annotators. All experiments use existing public or externally released research datasets under their documented access conditions. We do not attempt to re-identify participants, link records across datasets, or infer personal identities. The derived benchmark stores dataset-local subject or record identifiers only as technical locators for loading signal windows, and evaluation prompts avoid direct personal identifiers. The data used in this work consists of physiological signals and structured labels rather than free-form text, images, or audio, so offensive content risk is minimal.

\paragraph{Use of AI assistants.}
AI assistants, including ChatGPT, Claude, and Codex, were used for coding assistance, debugging, result summarization, language polishing, and checklist preparation. All scientific claims, experimental settings, numeric results, citations, and final writing decisions were checked and approved by the authors. AI assistants were not used to generate participant data, conduct human-subject interactions, or make clinical decisions.